# Fault diagnosis for open-circuit faults in NPC inverter based on knowledge-driven and data-driven approaches

Lei Kou, Chuang Liu , Guo-wei Cai, Jia-ning Zhou, Quan-de Yuan, Si-miao Pang

## Abstract

In this study, the open-circuit faults diagnosis and location issue of the neutral-point-clamped (NPC) inverters are analysed. A novel fault diagnosis approach based on knowledge driven and data driven was presented for the open-circuit faults in insulated-gate bipolar transistors (IGBTs) of NPC inverter, and Concordia transform (knowledge driven) and random forests (RFs) technique (data driven) are employed to improve the robustness performance of the fault diagnosis classifier. First, the fault feature data of AC in either normal state or open-circuit faults states of NPC inverter are analysed and extracted. Second, the Concordia transform is used to process the fault samples, and it has been verified that the slopes of current trajectories are not affected by different loads in this study, which can help the proposed method to reduce overdependence on fault data. Moreover, then the transformed fault samples are adopted to train the RFs fault diagnosis classifier, and the fault diagnosis results show that the classification accuracy and robustness performance of the fault diagnosis classifier are improved. Finally, the diagnosis results of online fault diagnosis experiments show that the proposed classifier can locate the open-circuit fault of IGBTs in NPC inverter under the conditions of different loads.

## 1 Introduction

Neutral-point-clamped (NPC) inverters have been successfully applied for renewable energy and industrial applications with the advantages of low harmonics, lower-voltage stress in power switches, high power factor and low switching losses [1, 2]. However, NPC inverters are composed of various devices. Among them, the power switches are one of the most fragile devices, thus play an important role to keep the reliability of the NPC inverter systems [3]. Hence, condition monitoring and fault diagnosis for NPC inverter are necessary to reduce the risks and economic losses caused by faults [4, 5].

Power device faults are usually divided into open-circuit faults and short-circuit faults [6]. The short-circuit faults in insulated-gate bipolar transistors (IGBTs) are destructive and usually protected by the standard hardware circuits [7, 8]. However, the open-circuit faults in IGBTs are usually caused by thermal cycling, excessive collector current and gate-driver fault, which do not cause serious damage, but may lead secondary faults to other devices when the power-electronic converters run with the open-circuit fault for a long time [9, 10]. Therefore, many open-circuit fault diagnosis methods have been developed by researchers in inverters [11]. A fault diagnosis and protection technique for the interior permanent-magnet synchronous motors based wavelet packet transform and artificial neural network (ANN) was presented in [12]. The open-circuit fault in a switching device can be detected by monitoring the gating signal and corresponding voltage across the switch, but it is an impractical method for NPC inverter system, which has many switching devices [13]. Moreover, it is also impossible to locate the fault switches by the conventional methods, which only use the distortion of outputs, and the reason is that the distortion of outputs is the same in some fault cases [14]. A fault detection method for the switch devices of three-parallel power conversions in a wind-turbine system was proposed in [15], which only used the measured three-phase currents without additional current and voltage sensors, and the method can diagnose an open-circuit fault switch in the converter by a neural network. An open-switch fault detection method and two tolerance controls based on space vector modulation were presented in [16], which did not need the additional device, except three current sensors. Among data-driven methods, ANN is a popular supervised learning algorithm, which has been adopted by many researchers to train the fault diagnosis classifier [17-19]. However, ANN is vulnerable to cause the overfitting, affect the generalisation ability and the diagnosis results. Random forests (RFs) are an effective tool in classification and prediction, which will not incur overfitting because it consists of large quantities of independent decision trees and training samples [20, 21].

According to Peng *et al.* [22], there is a lack of researchers around the world to model the power electronics converters under faults conditions. What is more, it is extremely difficult to do that under the open-circuit faults states. A diagnosis system was presented in [23], which based on the analysis of the current-vector trajectory under the Concordia frame and the instantaneous frequency of the current vector in fault mode. A method for the detection and identification of the open-circuit faults in transistor was presented in [24], which can obtain the specific pattern to identify the fault switch based on the symmetry of the image projection around the current trajectory mass centre. However, the most existing diagnosis approaches are poor performance in adaptive capacity and knowledge acquisition, because the majority of the fault diagnosis system is still using a single diagnosis method. Most researches focus on fault mode and effect analysis. Yet, they are not used for the detection of faults [25]. The knowledge and information about the fault behaviour of power electronics converter is important for the fault diagnosis system, and the fault features are easily affected by loads in many cases. Therefore, it is essential to integrate expert knowledge into the training of the fault diagnosis system.

The knowledge-driven approach mainly depends on the given knowledge in related fields [26]. Conversely, the data-driven approach can achieve strong classification ability from the given data by the machine-learning algorithms, which relies less on the domain knowledge [27]. A new adaptable method that integrates knowledge-driven and data-driven approaches for inferring occupant dynamics in building was presented in [28]. A hybrid knowledge-based and data-driven method was proposed in [29] to identify semantically similar concepts. A model-based and data-driven method was proposed in [30] to find residuals with good fault detection performance, which was illustrated by designing a diagnosis system to monitor the air path through an internal combustion engine. To reduce the difficulty of modelling, the signal-based description of the drive system was adopted in [31], and then the data-driven method was carried on dSPACE platform-based traction system. Knowledge-based method has strong knowledge performance and fault feature transformation capabilities, which can help the data-driven method to reduce the over-reliance of fault data [32]. The data-driven approach can complete models from the knowledge data, which can improve the initial activity models [33]. On the basis of the above research, a novel fault diagnosis approach based on knowledge driven and data driven is presented in this paper, and Concordia transform (knowledge driven) and RFs technique (data driven) are employed to improve the adaptive performance of the fault diagnosis classifier.

Several different fault diagnosis methods have been studied for power-electronic converters, but many methods depend on fault models, which are difficult to build [34]. Some diagnosis methods do not consider the cost of operating environment or some simple data-driven methods are difficult to adapt to different loads. We have gained great inspiration and experience from their researches. Meanwhile, the methods of knowledge driven and data driven based on different fields are also different because the knowledge in different fields are different. The contributions of this paper are summarised as follows:

(i) It has been verified that the slopes of current trajectories are not affected by different loads. Moreover, the knowledge-driven method can be used for feature transform and to help the data-driven methods to reduce the dependence on data.

(ii) Since only three-phase AC currents are selected as the input features for diagnosis, the directly measured signal is less, and meanwhile the proposed method can be more easily suitable for other three-phase power electronics conversion systems. The proposed method can adapt to different loads, which can reduce the collection of fault samples.

(iii) The time scale is considered in this paper, and the proposed online fault diagnosis system can achieve 200 diagnosis results per 20 ms. The final fault locations are determined by using 200 diagnosis results, which make the final diagnosis result more accurate.

Fig. 1 shows the diagram of the fault diagnosis system for NPC inverter. This paper is organised as follows: the fault features of phase currents are analysed when the open-circuit faults occur in NPC inverter, which are presented in Section 2. Section 3 describes the method of extracting fault feature based on the knowledge about the slopes of the current trajectory, and then a fault diagnosis classifier is trained by using RFs algorithm. Section 4 describes the fault diagnosis experiment and the experimental results with the proposed method. Moreover, some conclusions are discussed in Section 5.

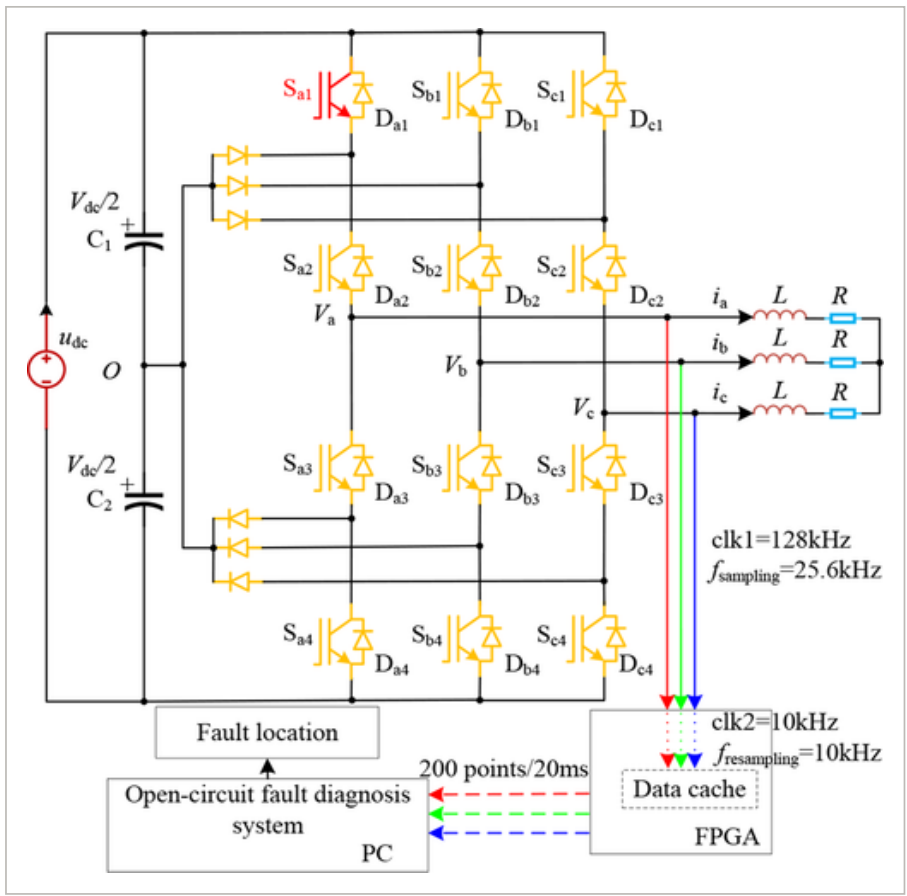

**Fig. 1**

Open in figure viewer | PowerPoint

*Fault diagnosis system for NPC inverter*

## 2 Fault features analysis of NPC inverter

The main circuit of NPC inverter is depicted in Fig. 1, the inverter has three arms and each arm has four IGBTs, and where $V_k$ is the output phase voltage and $i_k$ is the phase current, $k = a, b, c$. Here, $C1$ and $C2$ are the dc-link capacitances, respectively. Moreover, the open-circuit fault happened in IGBT $S_{a1}$. In this section, the fault features of phase currents are analysed when the open-circuit faults occur in IGBTs of NPC inverter.

The expression of output voltages $V_x$ can be described as

$$\begin{cases} V_a = \dfrac{V_{dc}}{2} S_a \\ V_b = \dfrac{V_{dc}}{2} S_b \\ V_c = \dfrac{V_{dc}}{2} S_c \end{cases} \quad (1)$$

where the switching functions $S_x$ are defined as

$$S_x = \begin{cases} 1, & x \text{ works in P mode}. \\ 0, & x \text{ works in O mode}. \\ -1, & x \text{ works in N mode}. \end{cases} \quad (2)$$

The fault features of phase currents are discussed with A phase and B-phase arms as examples. The system parameters are listed in Table 1. The open-circuit fault of IGBT is simulated by shutting down the control signal of IGBT.

**Table 1.** System parameters

| Parameters | Value |
| --- | --- |
| output frequency, Hz | 50 |
| dc voltage, V | 200 |
| switching frequency, kHz | 12.8 |
| load resistance, Ω | 10 |

Figs. 2 and 3 show the current flow paths according to work modes and phase current direction. The operating states of the IGBT and pole voltage in NPC inverter are listed in Table 2. Figs. 4 and 5 show the fault current waveform when the open-circuit faults occur in NPC inverter.

**Table 2.** Operating states of the switch and pole voltage

| Work modes | Switching states ($x = a$, $b$ and $c$) | | | | Pole voltage |
| --- | --- | --- | --- | --- | --- |
| | $S_{x1}$ | $S_{x2}$ | $S_{x3}$ | $S_{x4}$ | $V_{xO}$ |
| P | on | on | off | off | $V_{dc}/2$ |
| O | off | on | on | off | 0 |
| N | off | off | on | on | $-V_{dc}/2$ |

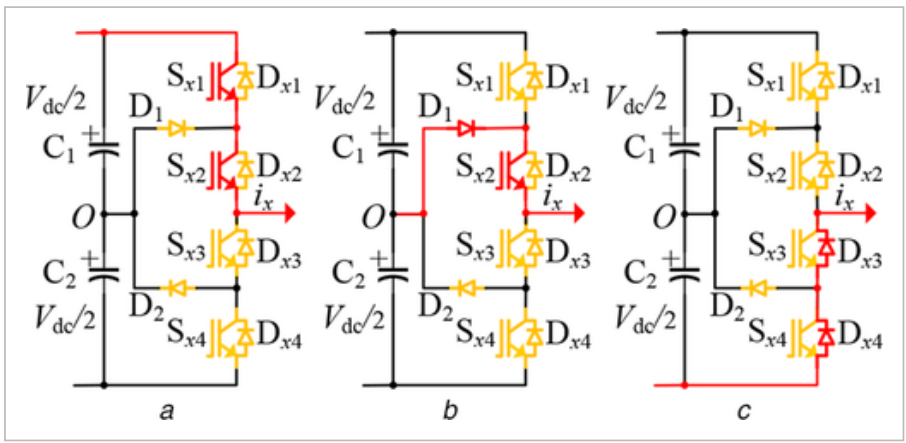

**Fig. 2**

Open in figure viewer | PowerPoint

Current paths according to switching state and current direction in an NPC inverter when $i_x > 0$

*(a)* Work modes P, *(b)* Work modes O, *(c)* Work modes N

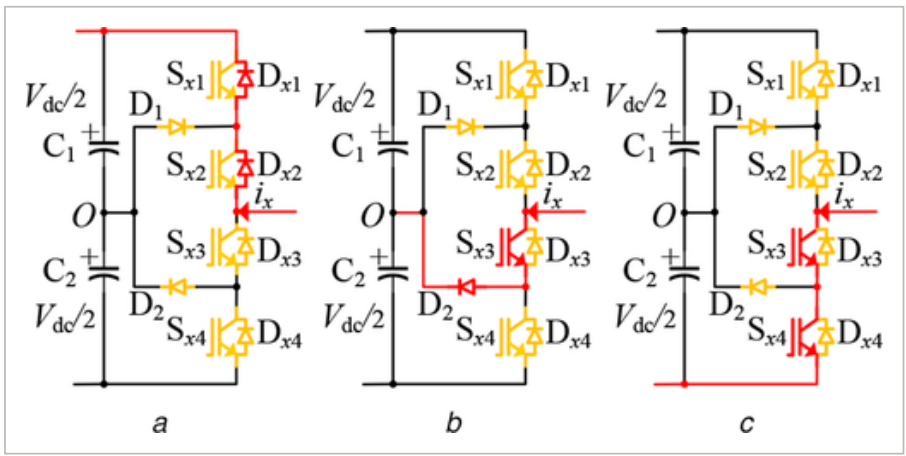

**Fig. 3**

Open in figure viewer | PowerPoint

Current paths according to switching state and current direction in an NPC inverter when $i_x < 0$

*(a)* Work modes P, *(b)* Work modes O, *(c)* Work modes N

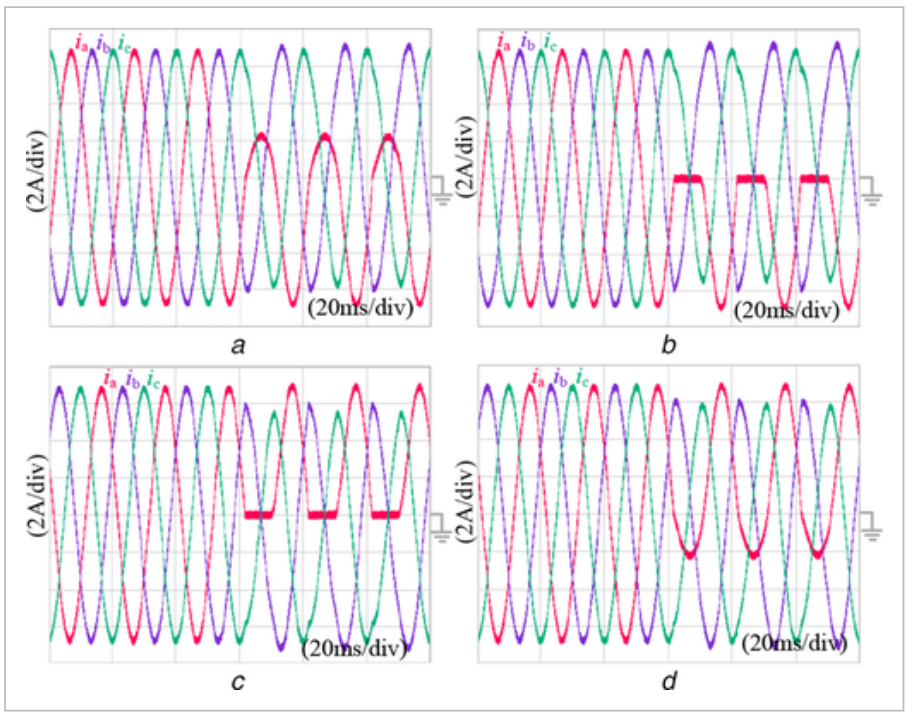

**Fig. 4**

A-phase fault current waveform

*(a)* $S_{a1}$ fault, *(b)* $S_{a2}$ fault, *(c)* $S_{a3}$ fault, *(d)* $S_{a4}$ fault

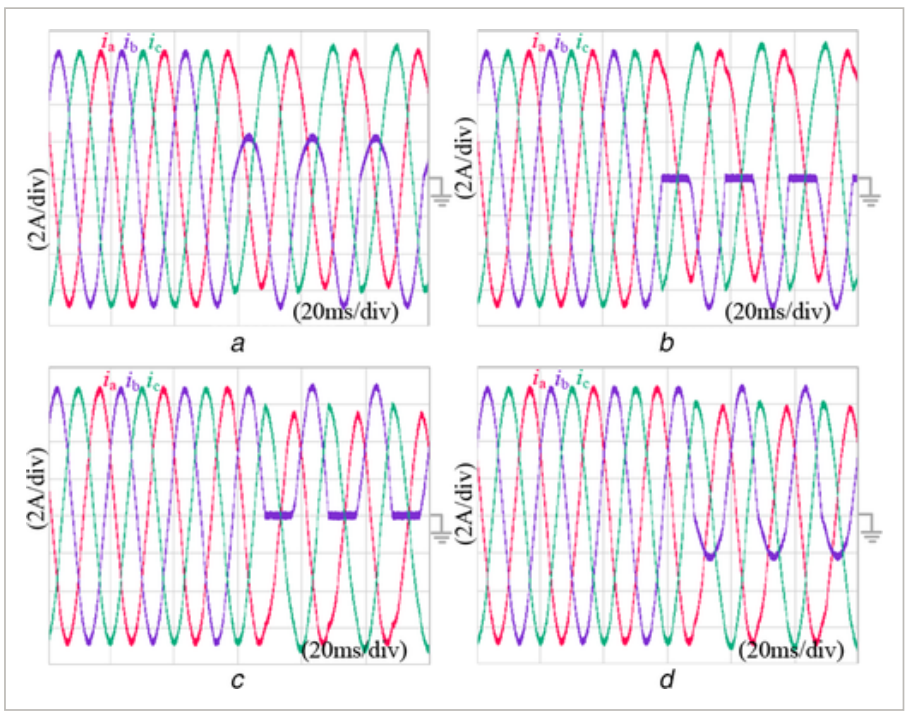

**Fig. 5**

B-phase fault current waveform

*(a)* $S_{b1}$ fault, *(b)* $S_{b2}$ fault, *(c)* $S_{b3}$ fault, *(d)* $S_{b4}$ fault

According to Figs. 2a, 4a and 5a, the work mode 'P' should be considered when the open-circuit fault occurs in IGBT $S_{x1}$. Compared to the positive output phase current $i_x$ in normal condition, the fault phase current is much smaller. According to Figs. 2b, 4b and 5b, the work modes 'P' and 'O' should be considered when the open-circuit fault occurs in IGBT $S_{x2}$, the positive output phase current $i_x$ is close to zero. According to Figs. 3c, 4c and 5c, the work modes 'O' and 'N' should be considered when the open-circuit fault occurs in IGBT $S_{x3}$, the negative output phase current $i_x$ is close to zero. According to Figs. 3c, 4d and 5d, the work mode 'N' should be considered when the open-circuit fault occurs in IGBT $S_{x4}$, the negative output phase current $i_x$ is much smaller than that in normal condition. The specific reasons about the above phenomena can be referred to [10], in which gives a very detailed explanation. According to Figs. 4 and 5, the three phases current change greatly when the fault happens, and different open-circuit fault locations will lead to different fault features of phase currents. Therefore, fault features of phase currents can be used for fault location.

## 3 Fault diagnosis classifier based on integrated knowledge-driven and data-driven approaches

In this section, the method based on the knowledge about the slopes of the current trajectory is described, and then the fault diagnosis classifiers trained by ANN and RFs algorithm are compared. Finally, the effectiveness of the proposed method is verified by robustness analysis.

### 3.1 Features processing by Concordia transform

To adapt to different loads, the three-phase currents of NPC inverter are processed by Concordia transformation, which can convert the three-phase system into a two-phase system. The slope of the current trajectory, obtained by Concordia transformation, can be maintained in a stable range. The Concordia transformation can be expressed as

$$\begin{bmatrix} i_\alpha \\ i_\beta \end{bmatrix} = m * \begin{bmatrix} 1 & -\frac{1}{2} & -\frac{1}{2} \\ 0 & \frac{\sqrt{3}}{2} & -\frac{\sqrt{3}}{2} \end{bmatrix} \begin{bmatrix} i_a \\ i_b \\ i_c \end{bmatrix} \quad (3)$$

where $m = \sqrt{2/3}$.

The three-phase AC currents of NPC inverters can be expressed as

$$\begin{cases} i_a = I_m \sin(\omega t) \\ i_b = I_m \sin\left(\omega t - \frac{2}{3}\pi\right) \\ i_c = I_m \sin\left(\omega t + \frac{2}{3}\pi\right) \end{cases} \quad (4)$$

where $I_m$ is the amplitude of three-phase AC currents and $i_a + i_b + i_c = 0$.
The slope of $\psi_1$ can be elaborated as

$$\psi_1 = \frac{i_\alpha(t)}{i_\beta(t)} = \frac{\sqrt{3}}{1 + 2(\sin(\omega t - (2/3)\pi)/\sin(\omega t))} \quad (5)$$

The slope of the current trajectory $\psi_2$ can be elaborated as

$$\psi_2 = \frac{i_\alpha(t) - i_\alpha(t-T)}{i_\beta(t) - i_\beta(t-T)}$$

$$= \frac{\sqrt{3}}{1 + 2(\sin(\omega t - (2/3)\pi) - \sin(\omega t - (2/3)\pi - T)/\sin(\omega t) - \sin(\omega t - T))}$$

(6)

According to formulas (5) and (6), $\psi_1$ and $\psi_2$ are only affected by $t$ and the sampling interval $T$. As shown in Fig. 6, $\psi_1$ and $\psi_2$ are only related to time $t$ and sampling interval $T$, but independent of amplitude $I_m$.

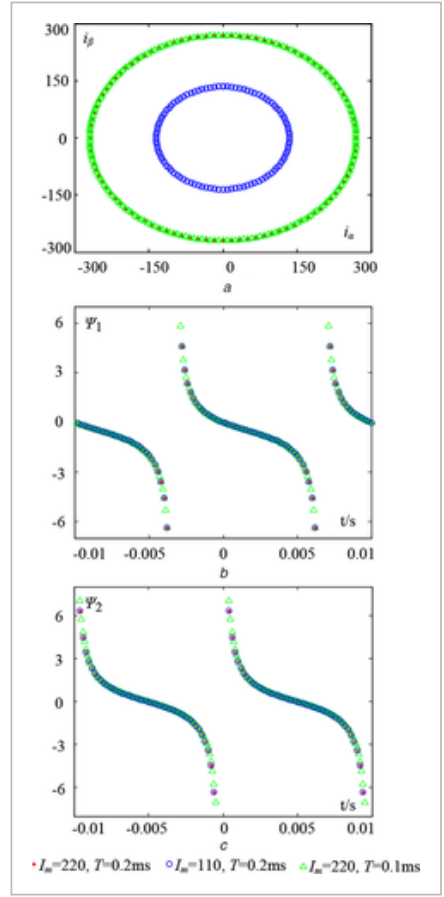

**Fig. 6**

Open in figure viewer | PowerPoint

*Current trajectories and slopes*

*(a)* Current trajectories, *(b)* $\psi_1$, *(c)* $\psi_2$

Assuming that $i_a + i_b + i_c = 0$, take $i_a$ as the first phase current, and then $i_{A\alpha}$ and $i_{A\beta}$ can be expressed as

$$\begin{cases} i_{A\alpha} = \sqrt{\frac{3}{2}} i_a \\ i_{A\beta} = \sqrt{2} i_b + \frac{i_a}{\sqrt{2}} \end{cases}$$

(7)

Moreover, then $\psi_{A1}$ and $\psi_{A2}$ can be expressed as

$$\begin{cases} \psi_{A1} = \dfrac{i_{A\alpha}(t)}{i_{A\beta}(t)} \\ \psi_{A2} = \dfrac{i_{A\alpha}(t) - i_{A\alpha}(t-T)}{i_{A\beta}(t) - i_{A\beta}(t-T)} \end{cases} \quad (8)$$

Take $i_b$ as the first phase current, then

$$\begin{cases} i_{B\alpha} = \sqrt{\dfrac{3}{2}}\, i_b \\ i_{B\beta} = \sqrt{2}\, i_c + \dfrac{i_b}{\sqrt{2}} \\ \psi_{B1} = \dfrac{i_{B\alpha}(t)}{i_{B\beta}(t)} \\ \psi_{B2} = \dfrac{i_{B\alpha}(t) - i_{B\alpha}(t-T)}{i_{B\beta}(t) - i_{B\beta}(t-T)} \end{cases} \quad (9)$$

Take $i_c$ as the first phase current, then

$$\begin{cases} i_{C\alpha} = \sqrt{\dfrac{3}{2}}\, i_c \\ i_{C\beta} = \sqrt{2}\, i_a + \dfrac{i_c}{\sqrt{2}} \\ \psi_{C1} = \dfrac{i_{C\alpha}(t)}{i_{C\beta}(t)} \\ \psi_{C2} = \dfrac{i_{C\alpha}(t) - i_{C\alpha}(t-T)}{i_{C\beta}(t) - i_{C\beta}(t-T)} \end{cases} \quad (10)$$

Figs. 7–11 show the features when the loads are $10\,\Omega$, and the loads in Figs. 12–16 are $20\,\Omega$.

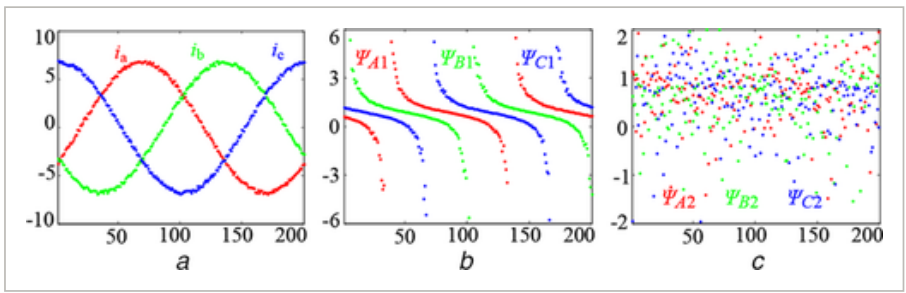

**Fig. 7**

Open in figure viewer | PowerPoint

*Features when a normal state with 10 Ω load*

*(a) $i_a$, $i_b$ and $i_c$, (b) $\Psi_{A1}$, $\Psi_{B1}$ and $\Psi_{C1}$, (c) $\Psi_{A2}$, $\Psi_{B2}$ and $\Psi_{C2}$*

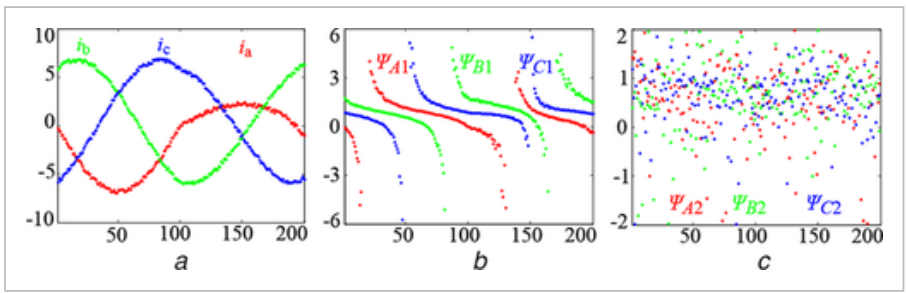

**Fig. 8**

Open in figure viewer | PowerPoint

*Features when $S_{a1}$ fault with 10 Ω load*

*(a) $i_a$, $i_b$ and $i_c$, (b) $\Psi_{A1}$, $\Psi_{B1}$ and $\Psi_{C1}$, (c) $\Psi_{A2}$, $\Psi_{B2}$ and $\Psi_{C2}$*

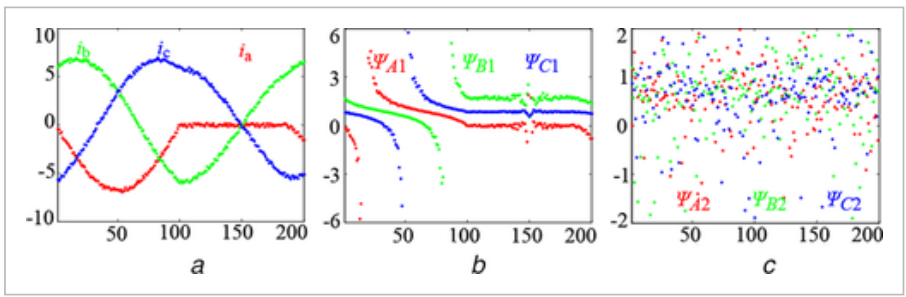

**Fig. 9**

Open in figure viewer | PowerPoint

*Features when $S_{a2}$ fault with 10 Ω load*

*(a) $i_a$, $i_b$ and $i_c$, (b) $\Psi_{A1}$, $\Psi_{B1}$ and $\Psi_{C1}$, (c) $\Psi_{A2}$, $\Psi_{B2}$ and $\Psi_{C2}$*

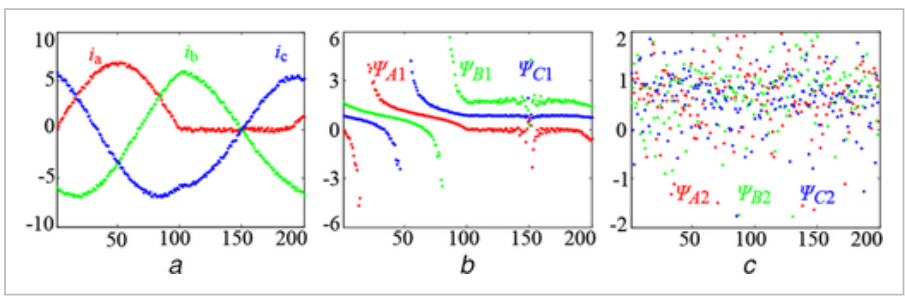

**Fig. 10**

Open in figure viewer | PowerPoint

*Features when $S_{a3}$ fault with 10 Ω load*

*(a) $i_a$, $i_b$ and $i_c$, (b) $\Psi_{A1}$, $\Psi_{B1}$ and $\Psi_{C1}$, (c) $\Psi_{A2}$, $\Psi_{B2}$ and $\Psi_{C2}$*

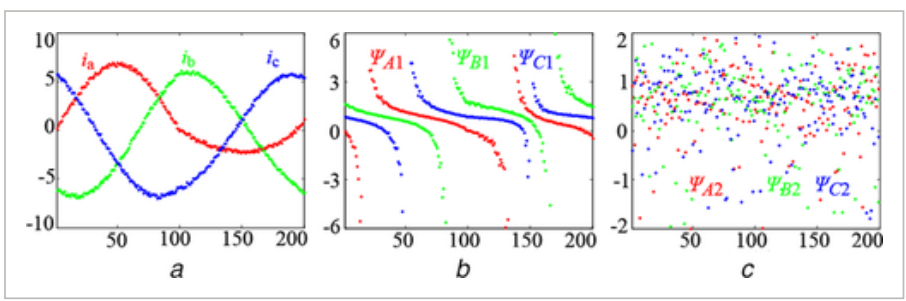

**Fig. 11**

Open in figure viewer | PowerPoint

*Features when $S_{a4}$ fault with 10 Ω load*

*(a) $i_a$, $i_b$ and $i_c$, (b) $\Psi_{A1}$, $\Psi_{B1}$ and $\Psi_{C1}$, (c) $\Psi_{A2}$, $\Psi_{B2}$ and $\Psi_{C2}$*

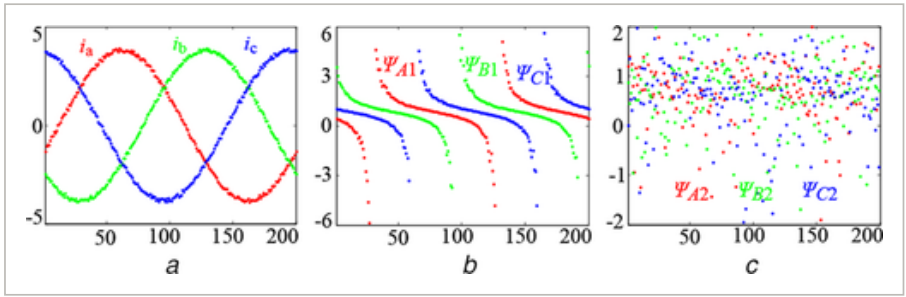

**Fig. 12**

Open in figure viewer | PowerPoint

*Features when a normal state with 20 Ω load*

*(a) $i_a$, $i_b$ and $i_c$, (b) $\Psi_{A1}$, $\Psi_{B1}$ and $\Psi_{C1}$, (c) $\Psi_{A2}$, $\Psi_{B2}$ and $\Psi_{C2}$*

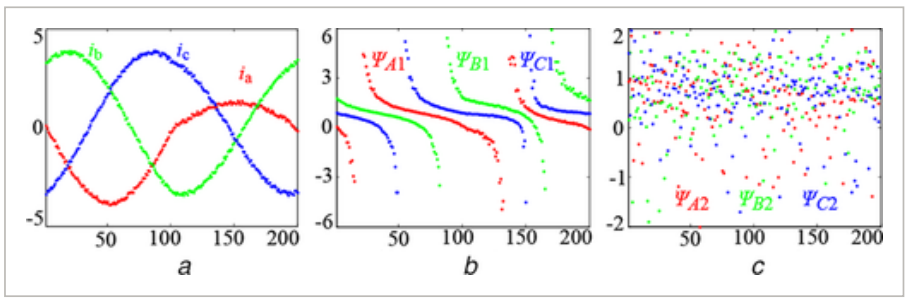

**Fig. 13**

Open in figure viewer | PowerPoint

*Features when $S_{a1}$ fault with 20 Ω load*

*(a) $i_a$, $i_b$ and $i_c$, (b) $\Psi_{A1}$, $\Psi_{B1}$ and $\Psi_{C1}$, (c) $\Psi_{A2}$, $\Psi_{B2}$ and $\Psi_{C2}$*

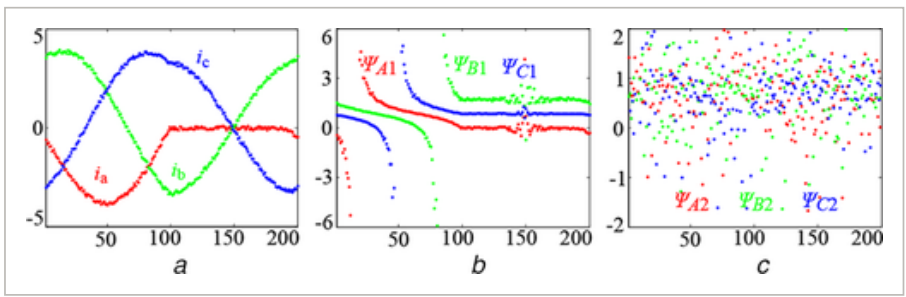

**Fig. 14**

Open in figure viewer | PowerPoint

*Features when $S_{a2}$ fault with 20 Ω load*

*(a) $i_a$, $i_b$ and $i_c$, (b) $\Psi_{A1}$, $\Psi_{B1}$ and $\Psi_{C1}$, (c) $\Psi_{A2}$, $\Psi_{B2}$ and $\Psi_{C2}$*

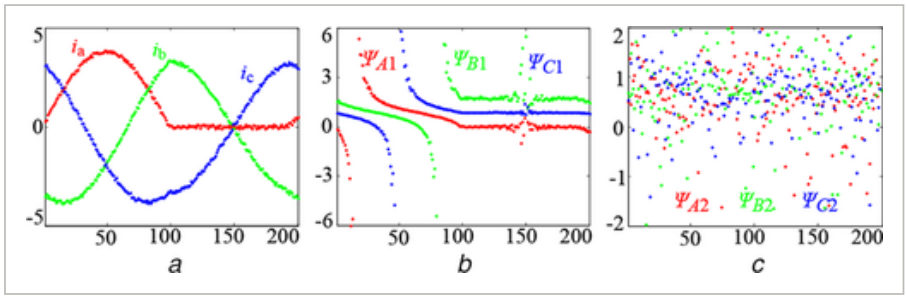

**Fig. 15**

Open in figure viewer | PowerPoint

*Features when $S_{a3}$ fault with 20 Ω load*

**(a)** $i_a$, $i_b$ and $i_c$, **(b)** $\Psi_{A1}$, $\Psi_{B1}$ and $\Psi_{C1}$, **(c)** $\Psi_{A2}$, $\Psi_{B2}$ and $\Psi_{C2}$

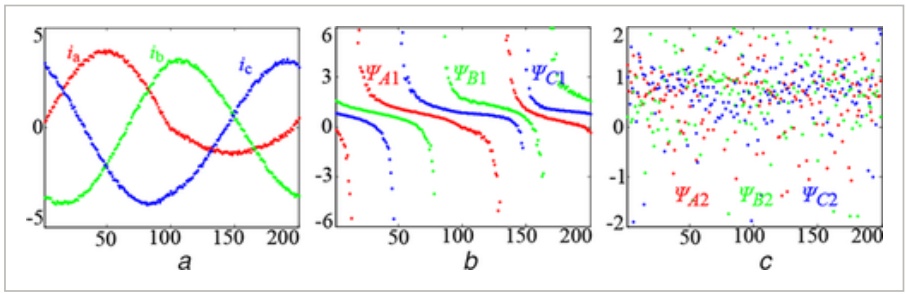

**Fig. 16**

Open in figure viewer | PowerPoint

*Features when $S_{a4}$ fault with 20 Ω load*

**(a)** $i_a$, $i_b$ and $i_c$, **(b)** $\Psi_{A1}$, $\Psi_{B1}$ and $\Psi_{C1}$, **(c)** $\Psi_{A2}$, $\Psi_{B2}$ and $\Psi_{C2}$

According to Figs. 7–16, different open-circuit fault locations will lead to different fault features. Since the magnitudes of the original fault currents are different under different loads, therefore, original fault currents have poor adaptability to loads. Compared to the original fault currents, the $(\Psi_{A1}, \Psi_{B1}, \Psi_{C1})$ and $(\Psi_{A2}, \Psi_{B2}, \Psi_{C2})$ have strong adaptive abilities under different loads. Therefore, as diagnosis features, the slopes of current trajectories are more suitable for open-circuit fault diagnosis of NPC inverters.

## 3.2 Training and evaluation of fault diagnosis classifier

RFs algorithm, which combines many decision trees, is an ensemble machine-learning algorithm for classification and prediction [35, 36]. RFs algorithm is commonly used to train the fault diagnosis classifier with the fault data samples. Like many of the supervised learning algorithms, the RFs algorithm employs the labelled samples to construct fault diagnosis classifier, so that the classifier can classify and discriminate the unlabelled samples. The difference is that RFs classifier is faster compared with other classifiers, especially for large datasets. Also, ANN algorithms are often adopted for fault diagnosis; therefore, the ANN algorithms are used for comparison, and the training process of ANN algorithms can refer to [18, 19]. To make the fault diagnosis classifier better adaptable to the different loads, the three-phase currents of NPC inverter will be processed by Concordia transformation, because the slopes of current trajectories have the ability to adapt to different loads.

Given $n$ labelled training samples $D = \{(x_1, y_1), \ldots, (x_n, y_n)\}$ as training data, where $x_i = \{i_a(t), i_b(t), i_c(t)\}$ and $y_i$ is the class labels (as shown in Table 3). RFs algorithm can make use of different training sets to increase the difference between classification models, which can improve the classification ability of classifier. Moreover, more detailed description about the training process of RFs can refer to [21].
Fig. 17 shows the flow of the data-driven approach with the original features, as shown in Fig. 18, the Concordia transformation is used to calculate $(\psi_{A1}, \psi_{B1}, \psi_{C1})$ and $(\psi_{A2}, \psi_{B2}, \psi_{C2})$ before the training of RFs. New representations of samples can be learnt by feature transformation, which can improve the performance of fault diagnosis classifier. The RFs model $\{h_1(X), h_2(X), \ldots, h_N(X)\}$ can be obtained from multiple decision trees model by training $N$ times. The final classification result is determined by a simple majority vote, which can refer to [36] and [37], and its expression can be elaborated as

$$H(X) = \arg\max_y \sum_{i=1}^{N} I(h_i(X) = y) \tag{11}$$

where $X$ is the input vector, $y$ is the output, $h_i(X)$ is the single decision tree model, $H(X)$ is the combined classification model and $I(\cdot)$ is the indicator function.

Table 3. Codes and labels of open-circuit faults

| Fault IGBT | Class labels y | | | | | | | | | | | |
|---|---|---|---|---|---|---|---|---|---|---|---|---|
| | d1 | d2 | d3 | d4 | d5 | d6 | d7 | d8 | d9 | d10 | d11 | d12 |
| normal state | 0 | 0 | 0 | 0 | 0 | 0 | 0 | 0 | 0 | 0 | 0 | 0 |
| $S_{a1}$ | 1 | 0 | 0 | 0 | 0 | 0 | 0 | 0 | 0 | 0 | 0 | 0 |
| $S_{a2}$ | 0 | 1 | 0 | 0 | 0 | 0 | 0 | 0 | 0 | 0 | 0 | 0 |
| $S_{a3}$ | 0 | 0 | 1 | 0 | 0 | 0 | 0 | 0 | 0 | 0 | 0 | 0 |
| $S_{a4}$ | 0 | 0 | 0 | 1 | 0 | 0 | 0 | 0 | 0 | 0 | 0 | 0 |
| $S_{b1}$ | 0 | 0 | 0 | 0 | 1 | 0 | 0 | 0 | 0 | 0 | 0 | 0 |
| $S_{b2}$ | 0 | 0 | 0 | 0 | 0 | 1 | 0 | 0 | 0 | 0 | 0 | 0 |
| $S_{b3}$ | 0 | 0 | 0 | 0 | 0 | 0 | 1 | 0 | 0 | 0 | 0 | 0 |
| $S_{b4}$ | 0 | 0 | 0 | 0 | 0 | 0 | 0 | 1 | 0 | 0 | 0 | 0 |
| ... | | | | | | ... | | | | | | |
| $S_{a1}$ and $S_{a3}$ | 1 | 0 | 1 | 0 | 0 | 0 | 0 | 0 | 0 | 0 | 0 | 0 |
| $S_{a2}$ and $S_{a3}$ | 0 | 1 | 1 | 0 | 0 | 0 | 0 | 0 | 0 | 0 | 0 | 0 |
| ... | | | | | | ... | | | | | | |

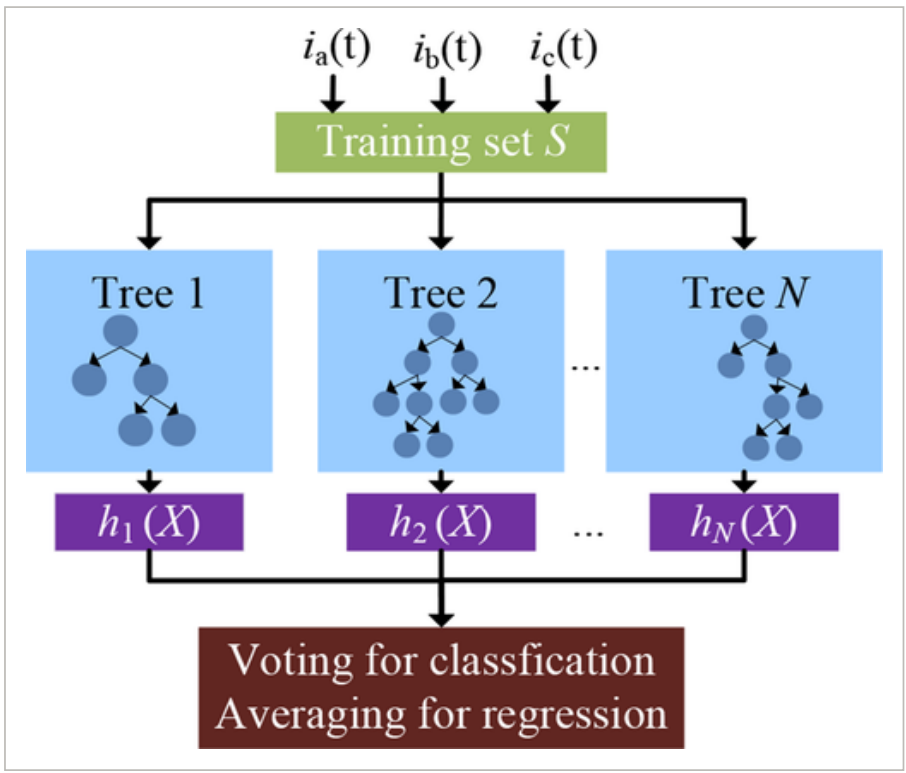

**Fig. 17**

Flowchart of the data-driven approach

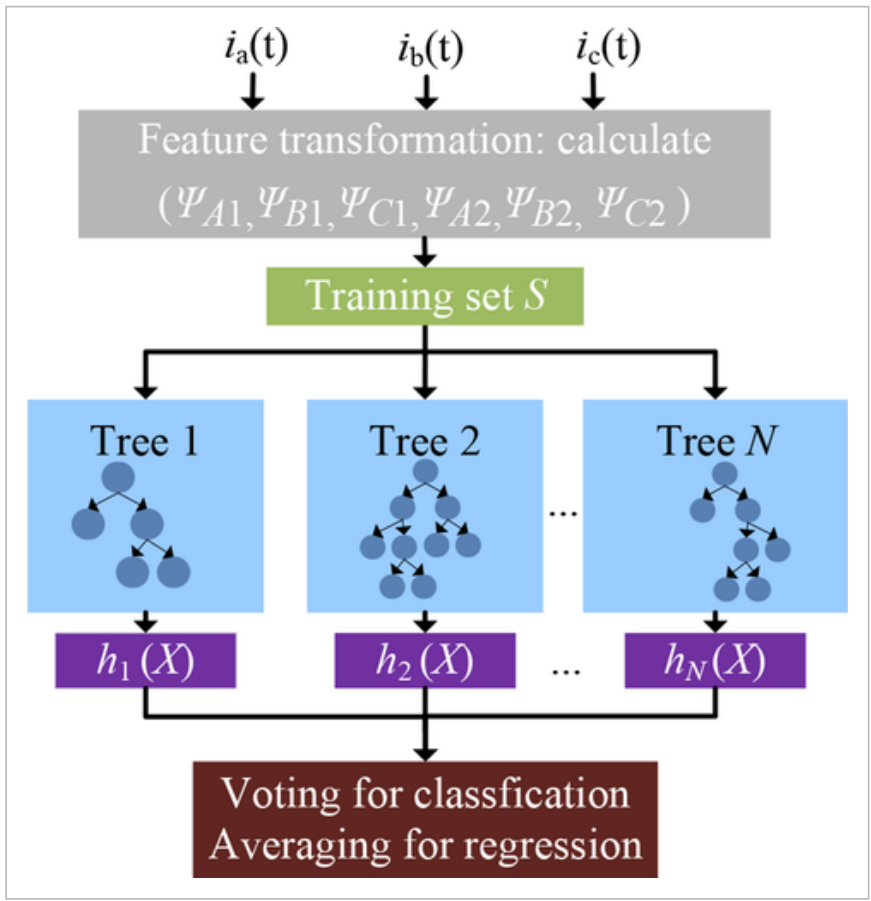

**Fig. 18**

*Flowchart of integrated knowledge-driven and data-driven approaches*

As shown in Table 3, when the open-circuit fault occurs in IGBT $S_{a1}$, $S_{a2}$, $S_{a3}$, $S_{a4}$, $S_{b1}$, $S_{b2}$, $S_{b3}$, $S_{b4}$, $S_{c1}$, $S_{c2}$, $S_{c3}$ and $S_{c4}$, the class labels $d_k(k = 1$ to $12) = 1$. On the basis of a great number of research experiences and attempts, each fault states consists of 21,000 fault samples, 70% of samples were adopted to train the RFs classifier, while the rest 30% of samples were used to evaluate the classifier. Moreover, the RFs are composed of 227 trees, where the highest accuracy is 0.9727 (as shown in Fig. 19).

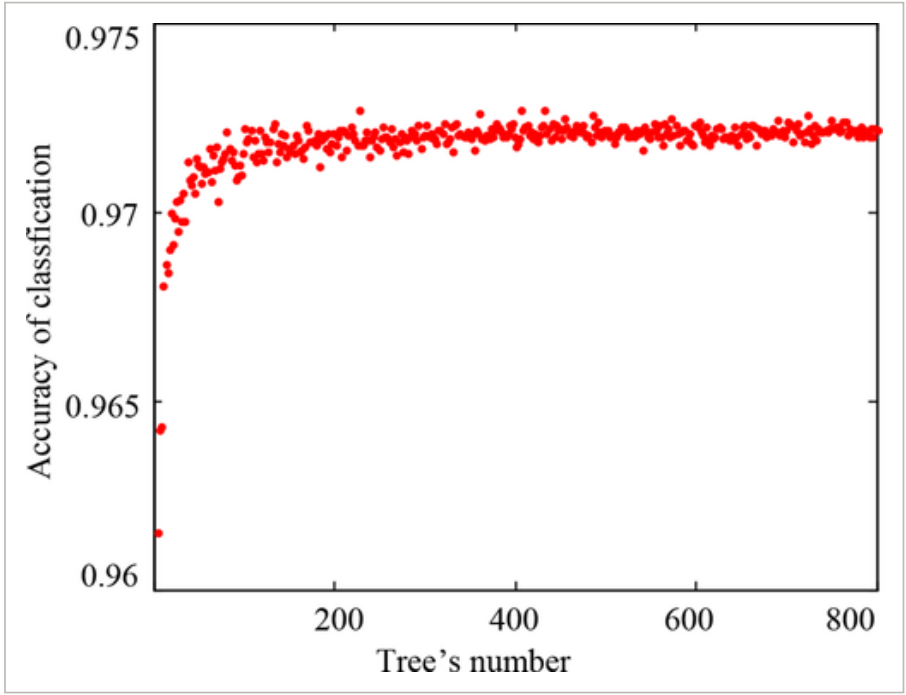

**Fig. 19**

Open in figure viewer | PowerPoint

*Influence of decision tree on performance in RFs*

After the training process, the fault diagnosis of the classifier is evaluated. Some samples and diagnosis results are shown in Table 4. To investigate the robustness of fault diagnosis classifier, which is based on integrated knowledge-based and data-driven approaches to different loads, the robustness analysis experiments are taken into consideration.

**Table 4.** Some samples and diagnosis results

| Fault IGBT | $i_a$ | $i_b$ | $i_c$ | Actual output y | Target output y |
|---|---|---|---|---|---|
| normal state | −3.603 | −3.169 | 6.772 | 000000000000 | 000000000000 |
| $S_{a1}$ | 1.619 | 3.189 | −4.808 | 100000000000 | 100000000000 |
| $S_{a1}$ | 1.557 | 3.323 | −4.881 | 100000000000 | 100000000000 |
| $S_{a2}$ | −4.669 | 1.588 | 3.081 | 010000000000 | 010000000000 |
| $S_{a2}$ | −4.733 | 1.543 | 3.190 | 010000000000 | 010000000000 |
| … | | | … | | |
| $S_{a1}$ and $S_{b2}$ | 3.105 | 0 | −3.105 | 101000000000 | 101000000000 |

| Fault IGBT | $i_a$ | $i_b$ | $i_c$ | Actual output $y$ | Target output $y$ |
|---|---|---|---|---|---|
| $S_{a1}$ and $S_{b2}$ | 3.335 | 0 | −3.335 | 101000000000 | 101000000000 |
| ... | | | | ... | |

Table 5 shows the performance of different fault diagnosis methods. Among all the fault diagnosis methods in Table 5, the training time of RFs with transformed features (the proposed method) is the shortest, that is to say, it has the fastest training speed. Moreover, meanwhile, the classification accuracy of the proposed method is the highest. According to Table 5, it can be seen that the classification accuracy of RFs is higher than that of ANN, and the training speed of RFs is faster than that of ANN.

**Table 5.** Comparison of performance

| Methods | Average diagnosis accuracy | Training time, min | Adaptability to fault data |
|---|---|---|---|
| ANN + original features | 0.9658 | 47.6 | weak |
| ANN + transformed features | 0.9672 | 38.2 | strong |
| RFs + original features | 0.9704 | 5.3 | weak |
| RFs + transformed features | 0.9727 | 4.6 | strong |

As shown in Table 6, Table 6 shows diagnosis accuracy of the traditional method. The fault diagnosis classifiers trained with the datasets when the load is $10\,\Omega$. The fault diagnosis classifier by the datasets of $10\,\Omega$ load does not apply to that of $20\,\Omega$ load. This method is too dependent on fault data of different loads.

**Table 6.** Diagnosis accuracy of the traditional method (RFs + original features) with fault samples of $10\,\Omega$

| Fault IGBT | Features $10\,\Omega$ | | Features $20\,\Omega$ | |
|---|---|---|---|---|
| | Diagnosis accuracy | Misdiagnosis rate | Diagnosis accuracy | Misdiagnosis rate |
| normal state | 0.9774 | 0.0226 | 0.5450 | 0.4550 |
| $S_{a1}$ | 0.9722 | 0.0278 | 0.4206 | 0.5794 |
| $S_{a2}$ | 0.9702 | 0.0298 | 0.5854 | 0.4146 |
| $S_{a3}$ | 0.9687 | 0.0313 | 0.5603 | 0.4397 |
| $S_{a4}$ | 0.9689 | 0.0311 | 0.5432 | 0.4568 |

| Fault IGBT | Features $10\,\Omega$ | | Features $20\,\Omega$ | |
| --- | --- | --- | --- | --- |
| | Diagnosis accuracy | Misdiagnosis rate | Diagnosis accuracy | Misdiagnosis rate |
| $S_{b3}$ | 0.9668 | 0.0332 | 0.5304 | 0.4696 |
| $S_{b4}$ | 0.9703 | 0.0297 | 0.7023 | 0.2977 |
| $S_{c1}$ | 0.9648 | 0.0352 | 0.5387 | 0.4613 |
| ⋮ | ⋮ | ⋮ | ⋮ | ⋮ |

As shown in Table 7, Table 7 shows the performance of the proposed fault diagnosis classifier with different loads. The robustness of the other datasets tell a similar story, so only some results on two datasets are given here, as follows. The fault diagnosis classifier here only uses datasets when the load is $10\,\Omega$, and the datasets of $10$ and $20\,\Omega$ are used to test. According to Tables 6 and 7, the results show that the proposed method is capable of being highly adaptive to different loads.

**Table 7.** Diagnosis accuracy of the proposed method (RFs + transformed features) with fault samples of $10\,\Omega$

| Fault IGBT | Features $10\,\Omega$ | | Features $20\,\Omega$ | |
| --- | --- | --- | --- | --- |
| | Diagnosis accuracy | Misdiagnosis rate | Diagnosis accuracy | Misdiagnosis rate |
| normal state | 0.9807 | 0.0193 | 0.9795 | 0.0205 |
| $S_{a1}$ | 0.9721 | 0.0279 | 0.9799 | 0.0201 |
| $S_{a2}$ | 0.9729 | 0.0271 | 0.9728 | 0.0272 |
| $S_{a3}$ | 0.9750 | 0.0250 | 0.9758 | 0.0242 |
| $S_{a4}$ | 0.9637 | 0.0363 | 0.9692 | 0.0308 |
| $S_{b1}$ | 0.9643 | 0.0357 | 0.9621 | 0.0379 |
| $S_{b2}$ | 0.9707 | 0.0293 | 0.9669 | 0.0331 |
| $S_{b3}$ | 0.9695 | 0.0305 | 0.9661 | 0.0339 |
| $S_{b4}$ | 0.9703 | 0.0297 | 0.9693 | 0.0307 |
| $S_{c1}$ | 0.9671 | 0.0329 | 0.9685 | 0.0315 |
| $S_{c2}$ | 0.9728 | 0.0272 | 0.9721 | 0.0279 |
| $S_{c3}$ | 0.9743 | 0.0257 | 0.9702 | 0.0298 |
| $S_{c4}$ | 0.9738 | 0.0262 | 0.9739 | 0.0261 |
| $S$ and $S$ | 0.9702 | 0.0298 | 0.9708 | 0.0292 |

# 4 Fault diagnosis experiments for NPC inverter

In this section, the fault diagnosis experiments for NPC inverter are detailed. The effectiveness of the proposed fault diagnosis classifier for detecting and classifying open-circuit faults of IGBTs is evaluated. The diagram of fault diagnosis system for NPC inverter is shown in Fig. 1, and the experimental setup is shown in Fig. 20. The bottom controller is the field-programmable gate array (EP4CE115F23) controller. The artificial intelligence (AI) fault diagnosis system is running on computer, the computer can be replaced by the embedded ARM chip (Cortex-A53) and the fault diagnosis system can be moved to the embedded Linux system environment. The bottom controller sends 200 three-phase AC currents values to the fault diagnosis system every 20 ms. The open-circuit faults of IGBTs in NPC inverter can be simulated by disabling the gate-driver signals of IGBTs. Figs. 21 and 22 show the fault diagnosis experiments and results when the open-circuit faults happen in $S_{a1}$ and ($S_{a1}$ and $S_{b2}$) under loads of 10 and $20\,\Omega$.

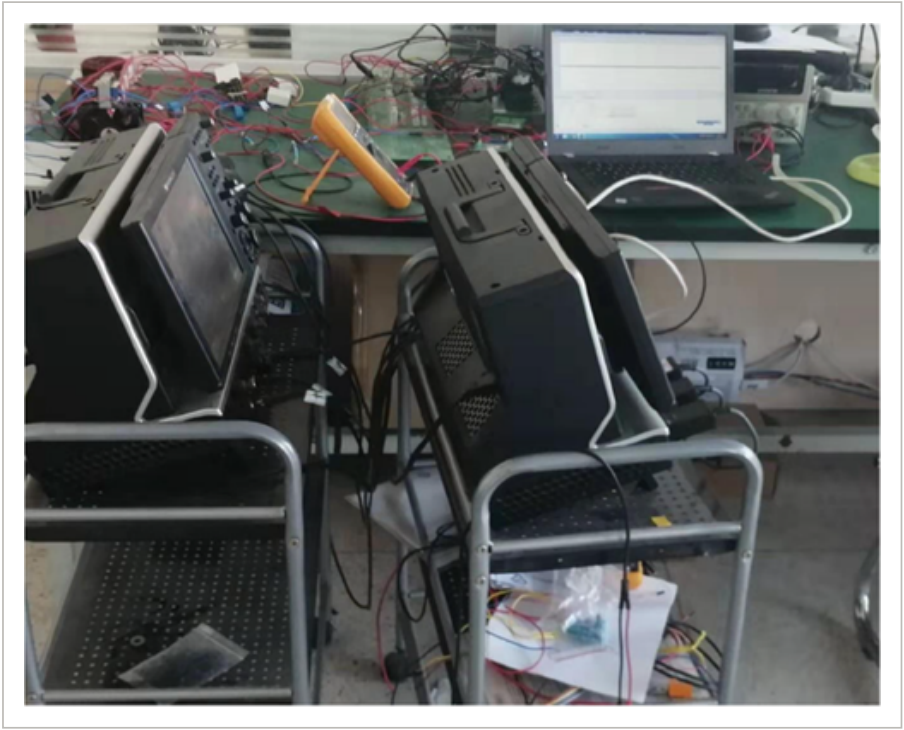

**Fig. 20**

Open in figure viewer | PowerPoint

*Experimental setup*

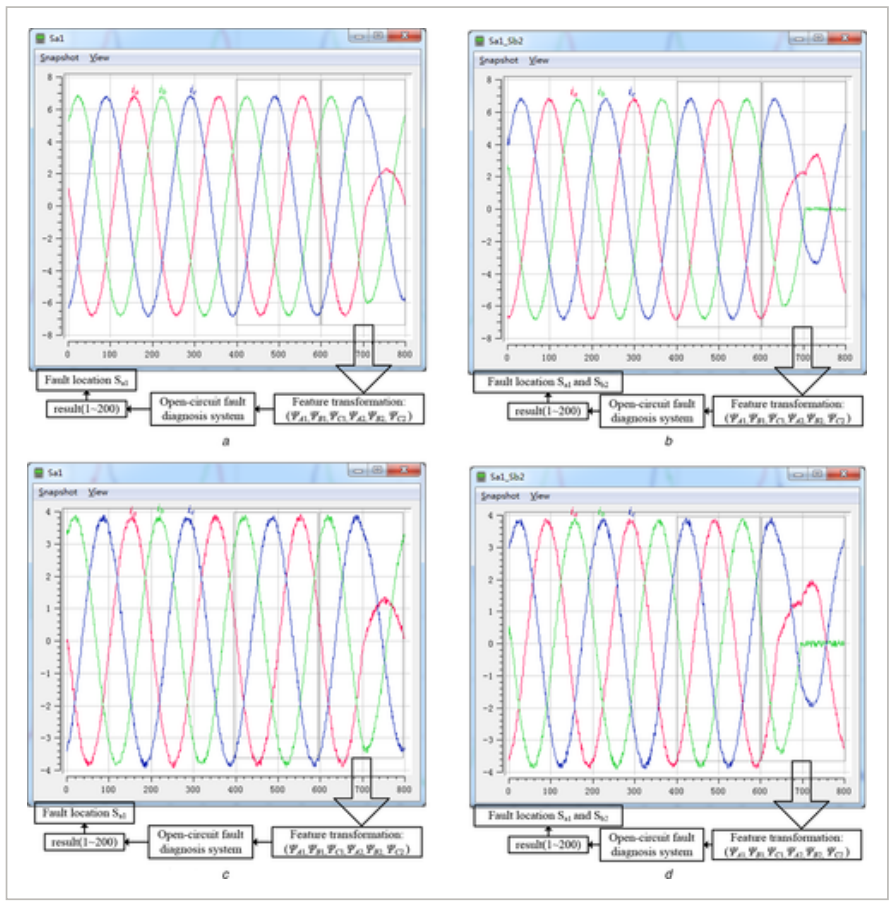

**Fig. 21**

*Fault diagnosis process*

*(a)*$S_{a1}$ fault with 10 Ω load, *(b)*$S_{a1}$ and $S_{b2}$ faults with 10 Ω load, *(c)*$S_{a1}$ fault with 20 Ω load, *(d)*$S_{a1}$ and $S_{b2}$ faults with 20 Ω load

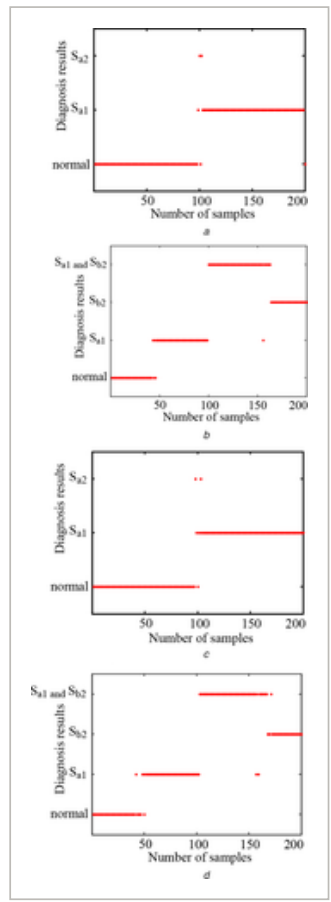

**Fig. 22**

Open in figure viewer | PowerPoint

*Fault diagnosis results*

*(a)* $S_{a1}$ fault with 10 Ω load, *(b)* $S_{a1}$ and $S_{b2}$ faults with 10 Ω load, *(c)* $S_{a1}$ fault with 20 Ω load, *(d)* $S_{a1}$ and $S_{b2}$ faults with 20 Ω load

As shown in Fig. 21, the hardware controller sends 200 sets of samples to the fault diagnosis system every 20 ms, and the fault diagnosis classifier can get 200 diagnosis results each time. Compared to time series, the instantaneous value is not affected by the time when the fault occurs, what is more, fault diagnosis classifier can also obtain more diagnosis results with fewer samples by using instantaneous values. According to Fig. 21, when the fault occurs, the faults features show about half a cycle, the faults locations can be located. Moreover, then sending out the protected signal and the fault location, therefore, the fault-tolerant processing or switching back-up equipment can be carried out in time. Thus, it can avoid more serious faults and economic losses.

As shown in Fig. 22, the diagnosis results in Fig. 22a show the case of open-circuit faults occur in IGBT $S_{a1}$ when the load is 10 Ω, and in the early stage of diagnosis results, 100 ahead of the results are in normal operation. The results between normal and fault are prone to misdiagnosis, and only a very small number of results were diagnosed as open-circuit faults in $S_{a2}$. Then, most of the remaining samples were diagnosed as open-circuit faults in $S_{a1}$. Therefore, the final fault location is $S_{a1}$. Fig. 22b shows the diagnosis results of open-circuit faults occur in IGBT $S_{a1}$ and $S_{b2}$ when the load is 10 Ω, the order of the diagnosis results is (normal) →($S_{a1}$ fault) →($S_{a1}$ and $S_{b2}$ fault)→($S_{b2}$ fault). Therefore, the final fault locations are $S_{a1}$ and $S_{b2}$. Similarly, it can also be applied to the case of 20 Ω load. The fault diagnosis classifier was trained with the fault samples when the load is 10 Ω, which is also applicable to 20 Ω. The experimental results prove that the proposed method can adapt to different loads, and it can be used to locate the open-circuit faults of IGBTs.

Since only three-phase AC currents are selected as the input features for diagnosis, so the directly measured signal is less, the proposed method can be more easily suitable for other three-phase power electronics conversion systems. However, the fault data samples of different converters or different working modes still need to be collected. If the proposed method can be implemented with the professional AI chip in the future, the real-time and performance of the method will be greatly improved.

# 5 Conclusion

This paper presents a novel fault diagnosis approach based on knowledge driven and data driven for open-circuit faults of IGBTs in NPC inverter. The knowledge-driven method can be used for feature transformation, and data-driven method is good at non-linear fitting. Combining the two methods can reduce the overdependence on fault data and make the method easy to implement.

The original fault features of three-phase AC currents and slopes of current trajectories are analysed, which found that the slopes of current trajectories are not affected by different loads. On the basis of this knowledge, the slopes dataset is adopted to train the fault diagnosis classifier with the data-driven method of RFs technique, and the proposed method has a high classification of fault locations. Compared to the existing data-driven methods, the proposed method can locate the fault IGBTs of NPC inverter under different loads, and it can adapt to different loads.

Finally, the online fault diagnosis experiments are carried out, and the results indicate the effectiveness of the proposed scheme. The final fault locations are determined by multi-group diagnosis results, which make the final diagnosis result more accurate.

# 6 Acknowledgment

This research is funded by the National Key R&D Program of China under Grant no. 2017YFB0903300.

# 7 References

1  Gupta A.K., and Khambadkone A.M.: 'A simple space vector PWM scheme to operate a three-level NPC inverter at high modulation index including overmodulation region, with neutral point balancing', *IEEE Trans. Ind. Appl.*, 2007, **43**, (3), pp. 751– 760
Crossref Web of Science® Google Scholar

2  Ceballos S., Pou J., and Robles E. *et al.*: 'Performance evaluation of fault-tolerant neutral-point-clamped converters', *IEEE Trans. Ind. Electron.*, 2010, **57**, (8), pp. 2709– 2718
Crossref Web of Science® Google Scholar

3  Yu Y., and Pei S.: 'Open-circuit fault diagnosis of neutral point clamped three-level inverter based on sparse representation', *IEEE Access*, 2018, **6**, pp. 73499– 73508
Crossref Web of Science® Google Scholar

4  Ge X., Pu J., and Gou B. *et al.*: 'An open-circuit fault diagnosis approach for single-phase three-level neutral-point-clamped converters', *IEEE Trans. Power Electron.*, 2018, **33**, (3), pp. 2559– 2570
Crossref Web of Science® Google Scholar

5  Ma K., Liserre M., and Blaabjerg F.: 'Operating and loading conditions of a three-level neutral-point-clamped wind power converter under various grid faults', *IEEE Trans. Ind. Appl.*, 2014, **50**, (1), pp. 520– 530
Crossref Web of Science® Google Scholar

6   Caseiro L.M.A., and Mendes A.M.S.:   'Real-time IGBT open-circuit fault diagnosis in three-level neutral-point-clamped voltage-source rectifiers based on instant voltage error' , *IEEE Trans. Ind. Electron.*, 2015, **62**, (3), pp. 1669– 1678
Crossref Web of Science® Google Scholar

7   Lamb J., and Mirafzal B.:   'Open-circuit IGBT fault detection and location isolation for cascaded multilevel converters' , *IEEE Trans. Ind. Electron.*, 2017, **64**, (6), pp. 4846– 4856
Crossref Web of Science® Google Scholar

8   Li X., Xu D., and Zhu H. *et al*.:   'Indirect IGBT over-current detection technique via gate voltage monitoring and analysis' , *IEEE Trans. Power Electron.*, 2019, **34**, (4), pp. 3615– 3622
Crossref Web of Science® Google Scholar

9   Lee J., Lee K., and Blaabjerg F.:   'Open-switch fault detection method of a back-to-back converter using NPC topology for wind turbine systems' , *IEEE Trans. Ind. Appl.*, 2015, **51**, (1), pp. 325– 335
Crossref Web of Science® Google Scholar

10   Choi U., Jeong H., and Lee K. *et al*.:   'Method for detecting an open-switch fault in a grid-connected NPC inverter system' , *IEEE Trans. Power Electron.*, 2012, **27**, (6), pp. 2726– 2739
Crossref Web of Science® Google Scholar

11   Lee J., and Lee K.:   'Open-switch fault tolerance control for a three-level NPC/T-type rectifier in wind turbine systems' , *IEEE Trans. Ind. Electron.*, 2015, **62**, (2), pp. 1012– 1021
Crossref Web of Science® Google Scholar

12   Khan M.A.S.K., and Rahman M.A.:   'Development and implementation of a novel fault diagnostic and protection technique for IPM motor drives' , *IEEE Trans. Ind. Electron.*, 2009, **56**, (1), pp. 85– 92
Crossref Web of Science® Google Scholar

13   Kim T., Lee W., and Hyun D.:   'Detection method for open-circuit fault in neutral-point-clamped inverter systems' , *IEEE Trans. Ind. Electron.*, 2009, **56**, (7), pp. 2754– 2763
Crossref Web of Science® Google Scholar

14   Choi U., Lee J., and Blaabjerg F. *et al*.:   'Open-circuit fault diagnosis and fault-tolerant control for a grid-connected NPC inverter' , *IEEE Trans. Power Electron.*, 2016, **31**, (10), pp. 7234– 7247
Web of Science® Google Scholar

15   Ko Y.J., Lee K.B., and Lee D.C. *et al*.:   'Fault diagnosis of three-parallel voltage-source converter for a high-power wind turbine' , *IET Power Electron.*, 2012, **5**, (7), pp. 1058– 1067
Crossref Web of Science® Google Scholar

16   Lee J., and Lee K.:   'An open-switch fault detection method and tolerance controls based on SVM in a grid-connected T-type rectifier with unity power factor' , *IEEE Trans. Ind. Electron.*, 2014, **61**, (12), pp. 7092– 7104
Crossref Web of Science® Google Scholar

17   Karmacharya I.M., and Gokaraju R.:   'Fault location in ungrounded photovoltaic system using wavelets and ANN' , *IEEE Trans. Power Deliv.*, 2018, **33**, (2), pp. 549– 559
Crossref Web of Science® Google Scholar

18 Abdullah A.: 'Ultrafast transmission line fault detection using a DWT-based ANN' , *IEEE Trans. Ind. Appl.*, 2018, **54**, (2), pp. 1182– 1193
Crossref Web of Science® Google Scholar

19 Khomfoi S., and Tolbert L.M.: 'Fault diagnostic system for a multilevel inverter using a neural network' , *IEEE Trans. Power Electron.*, 2007, **22**, (3), pp. 1062– 1069
Crossref Web of Science® Google Scholar

20 Sun J., Zhong G., and Dong J. *et al.*: 'Cooperative profit random forests with application in ocean front recognition' , *IEEE Access*, 2017, **5**, pp. 1398– 1408
Crossref Web of Science® Google Scholar

21 Shah A.M., and Bhalja B.R.: 'Fault discrimination scheme for power transformer using random forest technique' , *IET Gener. Transm. Distrib.*, 2016, **10**, (6), pp. 1431– 1439
Wiley Online Library Web of Science® Google Scholar

22 Peng T., Tao H.W., and Yang C. *et al.*: 'A uniform modeling method based on open-circuit faults analysis for NPC-three-level converter' , *IEEE Trans. Circuits Syst. II, Express Briefs*, 2019, **66**, (3), pp. 457– 461
Crossref Web of Science® Google Scholar

23 Peuget R., Courtine S., and Rognon J.: 'Fault detection and isolation on a PWM inverter by knowledge-based model' , *IEEE Trans. Ind. Appl.*, 1998, **34**, (6), pp. 1318– 1326
Crossref Web of Science® Google Scholar

24 Pires V.F., Amaral T.G., and Sousa D. *et al.*: ' Fault detection of voltage-source inverter using pattern recognition of the 3D current trajectory' . 2010 IEEE Region 8 Int. Conf. Computational Technologies in Electrical and Electronics Engineering (SIBIRCON), Listvyanka, 2010, pp. 617– 621
Crossref Google Scholar

25 Trunzer E., Weiß I., and Folmer J. *et al.*: ' Failure mode classification for control valves for supporting data-driven fault detection' . 2017 IEEE Int. Conf. Industrial Engineering and Engineering Management (IEEM), Singapore, 2017, pp. 2346– 2350
Crossref Google Scholar

26 Fan X.R., Kang M.Z., and Heuvelink E. *et al.*: 'A knowledge-and-data-driven modeling approach for simulating plant growth: a case study on tomato growth' , *Ecol. Model.*, 2015, **312**, pp. 363– 373
Crossref Web of Science® Google Scholar

27 Azkune G., Almeida A., and López-de-Ipiña D. *et al.*: 'Extending knowledge-driven activity models through data-driven learning techniques' , *Expert Syst. Appl.*, 2015, **42**, (6), pp. 3115– 3128
Crossref Web of Science® Google Scholar

28 Sonta A.J., Simmons P.E., and Jain R.K.: 'Understanding building occupant activities at scale: an integrated knowledge-based and data-driven approach' , *Adv. Eng. Inf.*, 2018, **37**, pp. 1– 13
Crossref Web of Science® Google Scholar

29 Pivovarov R., and Noémie E.: 'A hybrid knowledge-based and data-driven approach to identifying semantically similar concepts' , *J. Biomed. Inf.*, 2012, **45**, (3), pp. 471– 481
Crossref PubMed Web of Science® Google Scholar

30  Jung D., and Sundström C.: 'A combined data-driven and model-based residual selection algorithm for fault detection and isolation', *IEEE Trans. Control Syst. Technol.*, 2019, **27**, (2), pp. 616– 630
Crossref Web of Science® Google Scholar

31  Chen H., Jiang B., and Chen W. *et al*.: 'Data-driven detection and diagnosis of incipient faults in electrical drives of high-speed trains', *IEEE Trans. Ind. Electron.*, 2019, **66**, (6), pp. 4716– 4725
Crossref Web of Science® Google Scholar

32  Gao Z., Cecati C., and Ding S.X.: 'A survey of fault diagnosis and fault-tolerant techniques part ii: fault diagnosis with knowledge-based and hybrid/active approaches', *IEEE Trans. Ind. Electron.*, 2015, **62**, (6), pp. 3768– 3774
Crossref Web of Science® Google Scholar

33  Luo L.J., Bao S.Y., and Mao J.F. *et al*.: 'Industrial process monitoring based on knowledge data integrated sparse model and two-level deviation magnitude plots', *Ind. Eng. Chem. Res.*, 2018, **57**, (2), pp. 611– 622
Crossref CAS Web of Science® Google Scholar

34  Tariq M.F., Khan A.Q., and Abid M. *et al*.: 'Data-driven robust fault detection and isolation of three-phase induction motor', *IEEE Trans. Ind. Electron.*, 2019, **66**, (6), pp. 4707– 4715
Crossref Web of Science® Google Scholar

35  Su H., and Liu T.: 'Enhanced-online-random-forest model for static voltage stability assessment using wide-area measurements', *IEEE Trans. Power Syst.*, 2018, **33**, (6), pp. 6696– 6704
Crossref Web of Science® Google Scholar

36  Dai C., Liu Z., and Hu K. *et al*.: 'Fault diagnosis approach of traction transformers in high-speed railway combining kernel principal component analysis with random forest', *IET Electr. Syst. Transp.*, 2016, **6**, (3), pp. 202– 206
Wiley Online Library Web of Science® Google Scholar

37  Breiman L.: 'Random forests', *Mach. Learn.*, 2001, **45**, (1), pp. 5– 32
Crossref Web of Science® Google Scholar

## Citing Literature

**Citation Statements** beta

| Supporting | Mentioning | Contrasting |
|---|---|---|
| 0 | 16 | 0 |

Explore this article's citation statements on **scite.ai**

powered by **scite_**

**Number of times cited according to CrossRef:** 9

Badii Gmati, Imed Jlassi, Sejir Khojet El Khil, Antonio J. Marques Cardoso, Open-switch fault diagnosis in voltage source inverters of PMSM drives using predictive current errors and fuzzy logic approach, IET Power Electronics, 10.1049/pel2.12098, **14**, 6, (1059-1072), (2021).

Wiley Online Library

Tao Chen, Yuedou Pan, Zhanbo Xiong, Fault diagnosis scheme for single and simultaneous open-circuit faults of voltage-source inverters on the basis of fault online simulation, Journal of Power Electronics, 10.1007/s43236-020-00209-1, **21**, 2, (384-395), (2021).
Crossref

Tao Chen, Yuedou Pan, Current Vector Phase-Based Diagnostic Method for Multiple Open-Circuit Faults in Voltage-Source Inverters, IEEJ Transactions on Electrical and Electronic Engineering, 10.1002/tee.23412, **16**, 7, (1005-1012), (2021).
Wiley Online Library

Won-Jae Kim, Sang-Hoon Kim, ANN design of multiple open-switch fault diagnosis for three-phase PWM converters, IET Power Electronics, 10.1049/iet-pel.2020.0795, **13**, 19, (4490-4497), (2021).
Wiley Online Library

Lijuan Shi, Ang Li, Lei Zhang, Sustainable Fault Diagnosis of Imbalanced Text Mining for CTCS-3 Data Preprocessing, Sustainability, 10.3390/su13042155, **13**, 4, (2155), (2021).
Crossref

Shuiqing Xu, Juxing Wang, Mingyao Ma, Open-circuit fault diagnosis method for three-level neutral point clamped inverter based on instantaneous frequency of phase current, Energy Conversion and Economics, 10.1049/enc2.12021, **1**, 3, (264-271), (2020).
Wiley Online Library

Bindu Sharan, Tushar Jain, Spectral analysis-based fault diagnosis algorithm for 3-phase passive rectifiers in renewable energy systems, IET Power Electronics, 10.1049/iet-pel.2020.0510, **13**, 16, (3818-3829), (2020).
Wiley Online Library

Yunjun Yu, Xiaoming Li, Lili Wei, Fault Tolerant Control of Five-Level Inverter Based on Redundancy Space Vector Optimization and Topology Reconfigruation, IEEE Access, 10.1109/ACCESS.2020.3033805, **8**, (194342-194350), (2020).
Crossref

Tao Chen, Yuedou Pan, Similarity Analysis-Based Diagnosis Method for Open-Circuit Faults of Inverters in PMSM Drive Systems, IOP Conference Series: Earth and Environmental Science, 10.1088/1755-1315/571/1/012025, **571**, (012025), (2020).
Crossref